\pdfoutput=1
\documentclass[conference]{IEEEtran}
\IEEEoverridecommandlockouts
\usepackage{cite}
\usepackage{amsmath,amssymb,amsfonts}
\usepackage{graphicx}
\usepackage{textcomp}
\usepackage{xcolor}
\usepackage{algorithm}
\usepackage{algorithmic}

\def\BibTeX{{\rm B\kern-.05em{\sc i\kern-.025em b}\kern-.08em
    T\kern-.1667em\lower.7ex\hbox{E}\kern-.125emX}}
\begin{document}


\title{Controllable Narrative Rendering for Enhanced Assisted Writing}

\author{
	\IEEEauthorblockN{
		Mingzhe Lu\textsuperscript{1,2},
		Yanbing Liu\textsuperscript{1,2},
		Jiayue Wu\textsuperscript{1,2},
		Jiarui Zhang\textsuperscript{1,2},
		Qihao Wang\textsuperscript{1,2},
		Yue Hu\textsuperscript{1,2},
		Yunpeng Li\textsuperscript{1,2,*},
		Yangyan Xu\textsuperscript{3}
	}
	\vspace{1ex}
	\IEEEauthorblockA{\textsuperscript{1}Institute of Information Engineering, Chinese Academy of Sciences, Beijing, China}
	\IEEEauthorblockA{\textsuperscript{2}School of Cyber Security, University of Chinese Academy of Sciences, Beijing, China}
	\IEEEauthorblockA{\textsuperscript{3}HiThink Research}
	\IEEEauthorblockA{\textsuperscript{*}Corresponding author: \texttt{liyunpeng@iie.ac.cn}}
}

\maketitle

\begin{abstract}
	Despite the remarkable proficiency of large language models (LLMs) in basic writing assistance, their utility in creative writing is fundamentally hindered by a persistent binary failure. This issue manifests as an oscillation between safe, surface-level editing, referred to as remedial polishing, and destructive, uncontrolled plot expansion. This dilemma defines a critical trade-off between narrative fidelity and descriptive intensity.
	We propose \textsc{Loom}, an assisted writing framework grounded in the narratological distinction between story and discourse. Loom employs a three-layer pipeline that operationalizes an intent-centered semiotic chain-of-thought to enforce precise control over narrative intent and rendering density. This architecture separates the generation of perceptual material from syntactic insertion, ensuring that enhancement occurs without violating the original event structure.
	Our comprehensive evaluation, which includes LLM-based metrics and human assessment, demonstrates that Loom successfully resolves this fundamental tension. Loom achieves the highest overall quality score, yielding substantial gains in factual integrity and descriptive intensity compared to state-of-the-art baselines.
\end{abstract}


\section{Introduction}

Writing assistance technologies have advanced rapidly since the advent of large language models (LLMs), showing proficiency in grammatical correction, fluency enhancement, and stylistic rewriting~\cite{krishna2020reformulating,reif2022recipe}. In both general and professional contexts, these tools act as sophisticated editors, resolving linguistic friction and ensuring structural soundness. State-of-the-art models have largely solved the problem of textual correctness by mastering the foundational layers of writing, often described as ``faithfulness'' and ``fluency''.

Despite these capabilities, existing paradigms remain grounded in what we characterize as \textit{remedial polishing}. In this mode, systems focus on surface-level editing, polishing form, or rephrasing expressions to enhance clarity. However, creative writing demands a fundamental shift from strictly ``fixing'' text to actively shaping the reader's experience~\cite{zhang2024openpi2}, a process we define as \textit{narrative rendering}. Unlike remedial polishing, rendering seeks to transform the perceptual texture~\cite{genette1980narrative}, which encompasses atmosphere, mood, and sensory detail, without altering the underlying story events.

To illustrate this distinction, consider the narrative event: ``He walked into the room.'' A \textit{deep rendering} might depict the air as ``stale and heavy as if the walls had been holding their breath,'' invoking tension; alternatively, it might describe ``sunlight pooling along the floor,'' evoking warmth. Both renderings create strikingly different reading experiences while strictly preserving the same factual action.

However, achieving this specific level of control remains a formidable challenge. General-purpose models tasked with enhancement either default to safe, surface-level editing~\cite{dwivedi2024editeval}, which amounts to remedial polishing, or generate uncontrolled plot expansions when prompted for expressivity, often effectively hallucinating new events~\cite{ji2023survey}. Consequently, there remains no computational framework capable of precisely controlling enhancement within the perceptual layer without compromising the factual integrity of the source.

This limitation reflects a conflation of two distinct narratological layers: the \textit{story} (what happens) and the \textit{discourse} (how it is told). According to structuralist theory~\cite{genette1980narrative}, vivid narrative meaning emerges not just from adding details, but from the structured coordination of perceptual information and interpretive stance. Therefore, effective assistance requires a mechanism that decouples these layers and enables precise stylistic modification while preserving the event structure.

We propose \textsc{Loom}, an assisted writing framework for narrative rendering that preserves story events. Inspired by the narratological distinction between story and discourse , \textsc{Loom} employs a structured pipeline. This pipeline first uses the Perception Quota Layer to allocate constrained sensory material , which is then transformed into intent-aligned expressive functions by the Meaning Making Layer , before final integration via the Narrative Rendering Layer.

In summary, our contributions are as follows:
\begin{itemize}
	\item We propose \textsc{Loom}, a structured assisted writing framework for controllable narrative rendering. By employing a three-layer pipeline, \textsc{Loom} operationalizes the narratological semiotic chain-of-thought to enable precise control over both \textit{narrative intent} and \textit{rendering density}.
	\item We design a theoretically informed evaluation protocol grounded in the narratological distinction between story and discourse. Our multi-dimensional rubric explicitly measures expressive intensity and adherence to source events, addressing the limitations of standard metrics that conflate stylistic enrichment with plot alteration.
	\item We conduct comprehensive evaluations, including LLM-based metrics, human assessments, and ablation studies, demonstrating that \textsc{Loom} resolves the tension between factual integrity and descriptive intensity, yielding renderings more faithful to source events and better aligned with narrative intentions than strong LLM baselines.
\end{itemize}

\begin{figure*}[t]
	\centering
	\includegraphics[width=0.88\linewidth]{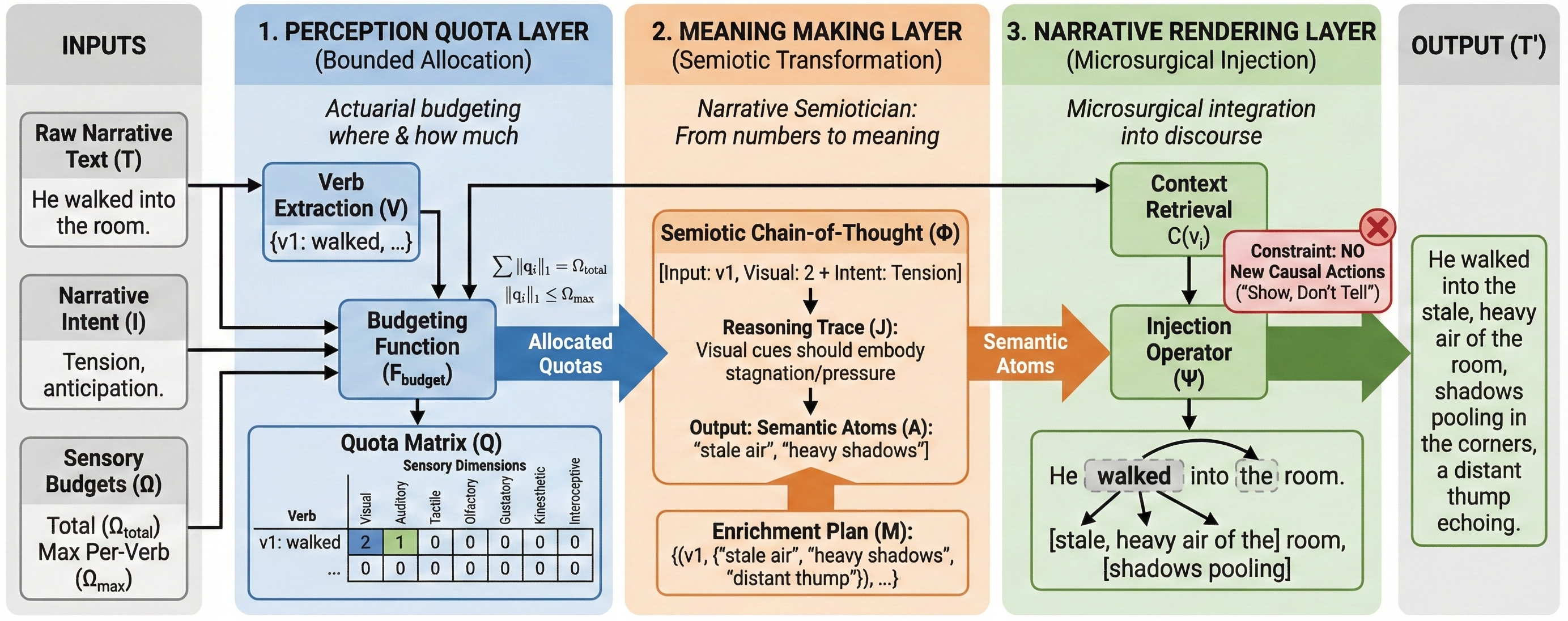}
	\caption{Overview of the \textsc{Loom} framework. The pipeline takes a raw narrative text, narrative intent, and density constraints as inputs. It operates through three stages: the Perception Quota Layer allocates sensory budgets based on intent; the Meaning Making Layer transforms abstract quotas into concrete semantic atoms via semiotic reasoning; and the Narrative Rendering Layer performs microsurgical injection to enrich the text while preserving the original event structure.}
	\label{fig:loom_overview}
\end{figure*}

\section{Related Work}

\textbf{Text Enrichment and Descriptive Expansion.}
Text enrichment has evolved from reconstruction losses for short premises~\cite{fan2018hierarchical,yao2019plan} to LLM-based methods targeting specific narrative qualities, such as psychology-grounded suspense planning~\cite{xie2024creating} and character-centric expansion~\cite{park2025character}. Stylistic frameworks like RSA-Control~\cite{wang2024rsa} employ pragmatic reasoning for tone shifts while anchoring semantics. However, these surface-level approaches lack a structural story-discourse distinction, often causing uncontrolled expansion that introduces new facts or alters pacing when seeking greater expressivity.

\textbf{Controllable and Planning-based Generation.}
Research has evolved from conditional training \cite{shirish2019ctrl,dathathri2019plug} to hierarchical planning, with Plan-and-Write \cite{yao2019plan} separating plot outlines from realization. Multi-agent frameworks like CritiCS \cite{bae2024collective} use collective critics for refinement, while DOC \cite{yang2023doc} applies detailed outline control. Agentic workflows employ self-correction mechanisms \cite{teleki2025survey}. Despite these advances, existing controls govern global attributes (genre, sentiment) or event sequences, lacking granularity to regulate \textit{perceptual density} within specific spans.

\section{Theoretical Foundation}
Compelling storytelling derives its power from a narratological mechanism progressing from perception to interpretation to discourse. Classical structuralist theory, notably Genette's work on focalization, posits that narrative meaning emerges from coordinating three distinct layers: perceptual information (what is shown), interpretive stance (how it is processed), and discourse rendering (how it is told). Cognitive narratology adds that readers experience stories not through raw description, but via perceptual cues selectively foregrounded for psychological significance.

Thus, textual vividness arises not from accumulating detail, but from a structured progression: framing perceptual cues, endowing them with thematic relevance, and shaping them into surface expression.

Mirroring this transmission, \textsc{Loom} operationalizes these stages: framing attention via \textit{Perception Quotation}, assigning significance via \textit{Meaning Making}, and enhancing expression via \textit{Narrative Rendering}. This isomorphism ensures that our pipeline is not merely a generative heuristic, but a rigorous operationalization of narrative craft.

\section{Method}
We introduce \textsc{Loom}, a structured agentic pipeline for controllable narrative rendering. As shown in Fig~\ref{fig:loom_overview}, it processes raw text $T$, intent $I$, budget $\Omega_{\text{total}}$, and limit $\Omega_{\text{max}}$ through three sequentially ordered layers: Perception Quota, Meaning Making, and Narrative Rendering, each employing a role-play-based semiotic chain-of-thought.

\subsection{Perception Quota Layer: Bounded Allocation}
The first stage functions as an actuarial budgeter. Its objective is to determine where to inject details and how much detail to inject, without yet generating specific content. This separation of volume from content is critical for controllability.

Let $V = \{v_1, v_2, \dots, v_n\}$ be the set of event-carrying verbs extracted from the input text $T$. We define a sensory space $\mathcal{S}$ consisting of seven dimensions corresponding to the taxonomy in Table I: Visual, Auditory, Olfactory, Gustatory, Tactile, Interoceptive, and Kinesthetic.

For each verb $v_i \in V$, the model must assign a quota vector $\mathbf{q}_i \in \mathbb{N}^7$, where each element $q_{i,j}$ represents the number of sensory atoms allocated to verb $v_i$ for sensory dimension $j$. The allocation is governed by an intent-driven budgeting function $F_{\text{budget}}$:

\begin{equation}
	\mathbf{Q} = F_{\text{budget}}(V, I, \Omega_{\text{total}}, \Omega_{\text{max}}),
\end{equation}

\noindent subject to two critical constraints:
\begin{equation}
	\text{s.t.} \quad \sum_{i=1}^{n} \|\mathbf{q}_i\|_1 = \Omega_{\text{total}}, \quad \|\mathbf{q}_i\|_1 \leq \Omega_{\text{max}}, \forall i \in \{1, \dots, n\}
\end{equation}

where $\|\mathbf{q}_i\|_1$ denotes the total sensory count allocated to verb $v_i$. The first constraint ensures that the system exactly exhausts the global budget without under-spending or over-spending. The second constraint prevents local inflation, ensuring that no single event is overwhelmed by excessive description. Verbs central to the narrative intent $I$ receive higher allocations within these bounds, while peripheral verbs receive zero quota.

\subsection{Meaning Making Layer: Semiotic Transformation}
The second stage functions as a narrative semiotician. It transforms the abstract numerical quotas from the previous layer into concrete, intent-aligned semantic symbols. This layer operates on the principle of atomic generation, producing minimal conceptual units (typically short phrases of 1--3 tokens representing concrete sensory impressions) rather than full sentences.

For every non-zero allocation $q_{i,j} > 0$ associated with verb $v_i$ and sensory dimension $s_j$, the model performs a semantic mapping $\Phi$. This process is driven by a semiotic chain-of-thought that explicitly reasons about how a physical sensation embodies the abstract intent $I$:

\begin{equation}
	(\mathcal{A}_{i,j}, \mathcal{J}_{i,j}) = \Phi(v_i, s_j, q_{i,j}, I).
\end{equation}

Here, $\mathcal{A}_{i,j} = \{a_1, \dots, a_k\}$ is a set of semantic atoms (e.g., ``stale air'' or ``distant thump'') serving as perceptual carriers. $\mathcal{J}_{i,j}$ represents the \textit{semiotic reasoning trace}. Far from being a mere justification, $\mathcal{J}_{i,j}$ encapsulates the core interpretative logic that imbues the physical atoms with specific narrative significance. It functions as the cognitive bridge between the concrete carrier and the abstract intent.

For the subsequent rendering stage, we aggregate all generated atoms for verb $v_i$ into a unified set $\mathcal{A}_i = \bigcup_{j} \mathcal{A}_{i,j}$, ensuring that the rendering layer receives a consolidated enrichment plan for each event.

\begin{table*}[h]
	\centering
	\caption{Sensory modalities and their representative semantic atoms.}
	\label{tab:sensory_atoms}
	\renewcommand{\arraystretch}{1.15}
	\begin{tabular}{l|c|c|c}
		\hline
		\textbf{Sensory Type} & \textbf{Common Semantic Atoms} &
		\textbf{Psychological Associations} & \textbf{Narrative Tendencies} \\
		\hline
		Visual & light, shadow, distance, color, shape &
		contrast $\rightarrow$ hope or pressure & symbolism, contrast \\
		Auditory & sound, echo, silence &
		noise $\rightarrow$ tension; silence $\rightarrow$ loneliness & rhythm, suspense \\
		Tactile & cold, heat, wet, dry, hard, soft &
		cold $\rightarrow$ isolation; heat $\rightarrow$ vitality & emotional shading \\
		Olfactory & fragrance, rot, freshness &
		clean scent $\rightarrow$ calm; decay $\rightarrow$ decline & evoking memory, atmosphere \\
		Gustatory & bitter, sweet, salty, sour &
		bitterness $\rightarrow$ setback; sweetness $\rightarrow$ relief & embodied experience \\
		Kinesthetic & speed, slowness, tremor, stillness &
		acceleration $\rightarrow$ tension; slowness $\rightarrow$ reflection & pacing control \\
		Interoceptive & heartbeat, breath, pressure &
		tightness $\rightarrow$ anxiety; warmth $\rightarrow$ reassurance & emotional immediacy \\
		\hline
	\end{tabular}
\end{table*}

\subsection{Narrative Rendering Layer: Microsurgical Injection}
The final stage executes a microsurgical integration of the semantic atoms into the original text. Unlike standard generative rewriting which often reshapes the entire paragraph, this layer applies a span-constrained injection operator.

Let $\text{span}(v_i)$ denote the start and end indices of verb $v_i$ in the original text $T$. The injection is performed sequentially following the linear order of verbs in $V$ to preserve temporal coherence. The rendering function $\Psi$ takes the local context $C(v_i)$ surrounding the verb and the generated semantic atoms $\mathcal{A}_i$, producing a refined local segment $T'_{i}$:

\begin{equation}
	T'_{i} = \Psi(C(v_i), \mathcal{A}_i, \text{Intent}(I)).
\end{equation}

The operator $\Psi$ is constrained by a strict ``Show, Don't Tell'' instruction and a negative constraint against adding new causal actions. The system identifies the precise narrative gap surrounding the verb $v_i$ and weaves the semantic atoms into the description. This localized approach ensures that the modification remains structurally subordinate to the original event backbone.

The logic of this incremental injection process is formalized in Algorithm~\ref{alg:injection}.

\begin{algorithm}[h]
	\caption{Microsurgical Narrative Injection}
	\label{alg:injection}
	\begin{algorithmic}[1]
		\REQUIRE $T$ (Source Text), $I$ (Intent), $\mathbb{M} = \{(v_i, \mathcal{A}_i)\}_{i=1}^{|V|}$ (Enrichment Plan)
		\ENSURE $T'$ (Rendered Text)
		\STATE $T' \leftarrow T$
		\FOR{\textbf{each} $(v_i, \mathcal{A}_i) \in \mathbb{M}$}
		\IF{$\mathcal{A}_i \neq \emptyset$}
		\STATE $\kappa \leftarrow C(v_i, T)$ \COMMENT{Extract local context window}
		\STATE $\tau \leftarrow \Psi(\kappa, \mathcal{A}_i, I)$ \COMMENT{Generate rendering via Eq. 4}
		\STATE $\sigma \leftarrow \text{span}(v_i, T')$ \COMMENT{Locate insertion coordinates}
		\STATE $T' \leftarrow \text{Update}(T', \sigma, \tau)$ \COMMENT{In-place injection}
		\ENDIF
		\ENDFOR
		\RETURN $T'$
	\end{algorithmic}
\end{algorithm}

The final output $T'$ is the concatenation of these locally refined segments, preserving the chronological and causal sequence of the original story $T$ while significantly enhancing its perceptual depth and atmospheric density.

\section{Experiments}
Evaluating narrative rendering is challenging, as it requires measuring stylistic intensity without penalizing plot preservation---a nuance conventional metrics miss. To the best of our knowledge, this work first formally defines and evaluates this task. Lacking established protocols, we propose a framework combining automatic metrics and human judgments to assess \textsc{Loom} on controllability, fidelity, and expressivity.

\subsection{Datasets}
We employ ROCStories~\cite{mostafazadeh2016corpus}, a collection of five-sentence narratives with strong causal backbones but minimal descriptive texture. This skeletal nature makes the dataset ideal for our task, as any sensory enrichment in the output can be attributed to the model's rendering process rather than the source text. Since \textsc{Loom} operates as an inference-only framework based on narratological prompting, we utilize the validation split for all experiments, requiring no additional training data.

\subsection{Evaluation Criteria}
Existing protocols, ranging from automatic metrics (e.g., BLEU, ROUGE) to human assessments of fluency and coherence, often conflate stylistic enhancement with plot alteration, inadvertently rewarding hallucinated events. 

To resolve this, we employ a unified rubric grounded in the \textit{story}-\textit{discourse} distinction~\cite{genette1980narrative}, which aligns rendered segments with the original structure and uses a flexible Likert scale~\cite{likert1932technique} (anchored at 1, 3, 5) to precisely decouple descriptive intensity from factual integrity.

\paragraph{\textbf{D1: Rendering Proportion Balance (RPB)}}
Evaluates the appropriateness of rendering volume and distribution.
\begin{itemize}
	\item \textbf{5} (Balanced): Rendering is moderate and distributed; no single sentence receives disproportionate elaboration.
	\item \textbf{3} (Locally Inflated): Global density is acceptable, but rendering is overly concentrated in a specific region.
	\item \textbf{1} (Overwhelming): Excessive rendering that effectively swallows the original narrative backbone.
\end{itemize}

\paragraph{\textbf{D2: Rendering Method Compliance (RMC)}}
Measures factual integrity, assessing if stylistic intent is achieved strictly through rendering rather than plot advancement.
\begin{itemize}
	\item \textbf{5} (Pure Rendering): No new events or causal relations introduced; effects arise strictly from texturing.
	\item \textbf{3} (Mixed Strategy): Stylistic effect achieved, but relies on micro-narrative extensions (minor added actions).
	\item \textbf{1} (Hallucination): Expands the story with new plot elements, violating boundary between story and discourse.
\end{itemize}

\paragraph{\textbf{D3: Rendering Stylistic Integration (RSI)}}
Measures the fluency and organic integration of inserted segments.
\begin{itemize}
	\item \textbf{5} (Seamless): Smoothly embedded; reads as a polished, coherent whole.
	\item \textbf{3} (Disjointed): Mechanically acceptable but exhibits transitional gaps requiring additional connectives.
	\item \textbf{1} (Abrupt): Unmotivated elements creating discontinuities that require causal justification.
\end{itemize}

\subsection{Baselines and Experimental Setup}
We compare \textsc{Loom} against a comprehensive suite of state-of-the-art models, broadly categorized into two paradigms. First, we evaluate standard language models using the Qwen2.5 family (7B, 14B, 32B) to verify scaling laws in open-weights performance. Second, we examine reasoning-enhanced (``thinking'') models, including Qwen3-235B, o4-mini, and GPT-5-thinking, to test whether intrinsic chain-of-thought capabilities can solve the narrative control problem.

Experiments were conducted on a randomly sampled set of 500 stories from the ROCStories test split. To ensure robustness, each story was rendered with three distinct narrative intents: psychological focus, environmental atmosphere, and mysterious tone. 

For evaluation, we employed claude-sonnet-4-thinking as the judge. The judge was provided with the full rubric defined in Section 4.3 and instructed to apply a ``strict and stingy'' scoring policy to prevent grade inflation. This rigorous setup ensures that high scores reflect genuine adherence to the deep rendering criteria.

\subsection{Result Analysis}
Table \ref{tab:baseline_dim_full} compares \textsc{Loom} with baselines across three dimensions, validating the binary failure modes discussed earlier and revealing significant divergence between standard generation and our structured pipeline.

The most striking disparity appears in Rendering Proportion Balance (RPB). All baselines, regardless of size or capability, scored near 1.0 (indicating excessive expansion), confirming that unconstrained LLMs interpret ``enrich'' as maximizing verbosity and obscuring the original narrative. In contrast, \textsc{Loom} achieves 2.83 (near the ideal 3), demonstrating how its perception quota layer prevents bloat through mathematical bounds on sensory injection.

\begin{table}[h]
	\centering
	\caption{LLM Assessment of Controllable Narrative Rendering (RPB, RMC, RSI).}
	\label{tab:baseline_dim_full}

	\begin{tabular}{l|ccc|c}
		\hline
		\textbf{LLM Eval} & \textbf{RPB} & \textbf{RMC} & \textbf{RSI} & \textbf{Average} \\
		\hline
		\multicolumn{5}{l}{\textbf{Language Model}} \\
		\hline
		Qwen2.5-7B 	 & 1.04 & 2.45 & 3.39 & 2.29 \\
		Qwen2.5-14B 	& 1.01 & 2.79 & 3.55 & 2.45 \\
		Qwen2.5-32B 	& 1.02 & 3.48 & 3.84 & 2.78 \\
		\hline
		\multicolumn{5}{l}{\textbf{Thinking Model}} \\
		\hline
		Qwen3-235B 	 & 1.02 & 3.78 & 3.96 & 2.92 \\
		o4-mini 	 & 1.04 & 3.96 & 4.31 & 3.10 \\
		GPT-5-thinking 	 & 1.02 & 4.29 & \textbf{4.33} & 3.21 \\
		\hline
		Loom(Ours) 	 & \textbf{2.83} & \textbf{4.83} & 3.93 & \textbf{3.86} \\
		\hline
	\end{tabular}
\end{table}

For Rendering Method Compliance (RMC), measuring hallucination absence, \textsc{Loom} achieves near-perfect 4.83. While GPT-5-thinking performs reasonably (4.29), smaller models struggle (Qwen2.5-7B: 2.45), often hallucinating events. \textsc{Loom}'s superiority stems from its meaning making layer, which structurally isolates stylistic enrichment from plot progression by generating atomic semantic units.

Regarding Stylistic Integration (RSI), GPT-5-thinking (4.33) slightly outperforms \textsc{Loom} (3.93), but this advantage comes from rewriting paragraphs, a strategy that sacrifices RMC and RPB for surface fluency. This reflects a common pattern where models generate fluent but uncontrolled expansions that drift from the original intent.

Ultimately, \textsc{Loom} achieves the superior trade-off: despite minor fluency gaps, it secures substantial narrative control gains, yielding the highest overall score (3.86) across all metrics.

\subsection{Human Evaluation}
To corroborate our automated findings, we conducted a fine-grained human evaluation involving three expert annotators. As shown in Table \ref{tab:human_eval_results}, \textsc{Loom} achieves a superior overall average of 3.77, significantly outperforming the GPT-5-thinking baseline (2.77). 

\begin{table}[h]
	\centering
	\caption{Human Evaluation: LOOM vs. GPT-5-thinking Baseline Across Three Dimensions.}
	\label{tab:human_eval_results}

	\begin{tabular}{l|ccc|c}
		\hline
		\textbf{Human Eval} & \textbf{RPB} & \textbf{RMC} & \textbf{RSI} & \textbf{Average}       \\
		\hline
		GPT-5-thinking  & 1.91 & 2.83 & 3.47 & 2.77   \\
		Loom(Ours)      & \textbf{3.46} & \textbf{4.20} & \textbf{3.64} & \textbf{3.77}    \\
		\hline
	\end{tabular}
\end{table}

Human judges consistently rated \textsc{Loom}'s density (RPB: 3.46) near the ideal moderate level, whereas the baseline's low score (1.91) confirms a systemic tendency toward excessive, uncontrolled expansion. Most critically, \textsc{Loom} maintained high factual integrity (RMC: 4.20), while the baseline was heavily penalized (2.83) for introducing hallucinations and plot deviations. These results confirm that our structural pipeline effectively shields the narrative backbone from distortion, breaking the tension that plagues standard generative models.

\subsection{Distance between Humans and Models}
To formally validate the reliability of our automated judge, we quantify the alignment between human ($H$) and model ($M$) evaluations. Adopting the distance metric proposed in prior work \cite{gado2025vist}, we first calculate the absolute differences for each dimension ($RPB, RMC, RSI$) as follows:

\begin{equation}
	\left\{
	\begin{aligned}
		d^{RPB}_{HM} &= |S^{RPB}_{H} - S^{RPB}_{M}| \\
		d^{RMC}_{HM} &= |S^{RMC}_{H} - S^{RMC}_{M}| \\
		d^{RSI}_{HM} &= |S^{RSI}_{H} - S^{RSI}_{M}|
	\end{aligned}
	\right.
\end{equation}
where $S_{H}$ and $S_{M}$ denote human and LLM judge scores. Finally, we compute the aggregate distance $d_{HM}$ as the mean of these component distances:

\begin{equation}
	d_{HM} = \frac{1}{3} \left( d^{RPB}_{HM} + d^{RMC}_{HM} + d^{RSI}_{HM} \right).
\end{equation}

Based on this metric, the calculated distance for \textsc{Loom} is 0.52, indicating a high degree of alignment given the subjective nature of the task. In contrast, the distance for the GPT-5-thinking baseline increases to 1.07. This divergence is primarily driven by the RMC dimension, where human judges were significantly stricter in penalizing hallucinations than the automated judge (Human: 2.83 vs. Model: 4.29). 

These results suggest that while our automated protocol is generally reliable, human experts remain indispensable for detecting subtle plot alterations. This underscores the broader reality that both generating and detecting hallucinations remain persistent challenges for large language models. Crucially, however, the relative ranking remains consistent across both evaluation modalities, confirming that the automated judge serves as a valid proxy for comparative assessment.

\section{Ablation Study}

To validate the necessity of each component in the \textsc{Loom} architecture, we systematically removed individual layers while keeping inputs constant, identifying the specific contributions of Meaning Making (MM) to semantic compliance and Narrative Rendering (NR) to structural preservation.

\begin{table}[h]
	\centering
	\caption{Effect of removing Meaning Making (MM).}
	\label{tab:ablation_mm}
	\begin{tabular}{l|ccc|c}
		\hline
		\textbf{Human Eval} & \textbf{RPB} & \textbf{RMC} & \textbf{RSI} & \textbf{Average} \\
		\hline
		\textsc{Loom} w/o MM & \textbf{3.47} & 3.86 & \textbf{3.78} & 3.70 \\
		\textsc{Loom} (Full) & 3.46 & \textbf{4.20} & 3.64 & \textbf{3.77} \\
		\hline
	\end{tabular}
\end{table}

\textbf{Effect of Meaning Making (MM).} Bypassing the MM layer forces the system to render raw sensory cues without intermediate semantic refinement. As shown in Table \ref{tab:ablation_mm}, this leads to a significant drop in Rendering Method Compliance (RMC, $4.20 \to 3.86$). Without intent-aligned semantic atoms, the model struggles to distinguish rendering from general rewriting. High Stylistic Integration ($3.78$) indicates that while the output remains fluent, it fails to adhere to the strict fidelity constraints.

\begin{table}[h]
	\centering
	\caption{Effect of removing Narrative Rendering (NR).}
	\label{tab:ablation_nr}
	\begin{tabular}{l|ccc|c}
		\hline
		\textbf{Human Eval} & \textbf{RPB} & \textbf{RMC} & \textbf{RSI} & \textbf{Average} \\
		\hline
		\textsc{Loom} w/o NR & \textbf{3.52} & 4.16 & 3.22 & 3.63 \\
		\textsc{Loom} (Full) & 3.46 & \textbf{4.20} & \textbf{3.64} & \textbf{3.77} \\
		\hline
	\end{tabular}
\end{table}

\textbf{Effect of Narrative Rendering (NR).} Replacing the microsurgical injection operator with a generative rewrite severely impacts textual coherence. Table \ref{tab:ablation_nr} shows a sharp decline in Stylistic Integration (RSI, $3.64 \to 3.22$). Without span injection, the model creates disjointed transitions and alters event structure, evidenced by RMC drop.

\begin{table}[h]
	\centering
	\caption{Performance with Human-in-the-Loop.}
	\label{tab:human_loop}
	\renewcommand{\arraystretch}{1.15}
	\begin{tabular}{l|ccc|c}
		\hline
		\textbf{Human Eval} & \textbf{RPB} & \textbf{RMC} & \textbf{RSI} & \textbf{Average} \\
		\hline
		\textsc{Loom} (Full) & 3.46 & \textbf{4.20} & 3.64 & 3.77 \\
		\textsc{Loom} + Human & \textbf{3.51} & 4.17 & \textbf{3.89} & \textbf{3.86} \\
		\hline
	\end{tabular}
\end{table}

\textbf{Human-in-the-Loop Optimization.} While fully automated, \textsc{Loom}'s layered design supports user intervention. As shown in Table \ref{tab:human_loop}, minimal human refinement of quotas or semantic atoms consistently improves performance across all dimensions ($3.77 \to 3.86$), demonstrating the framework's viability as a collaborative writing tool.

\begin{table}[h]
	\centering
	\caption{Control accuracy under varying \textbf{Total Quota} ($\Omega_{\text{total}}$).}
	\label{tab:quota_total}
	\renewcommand{\arraystretch}{1.15}
	\setlength{\tabcolsep}{3.5pt}
	\begin{tabular}{l|ccccccc}
		\hline
		\textbf{Quota Level} & \textbf{1} & \textbf{2} & \textbf{3} & \textbf{4} & \textbf{5} & \textbf{6} & \textbf{7} \\
		\hline
		Accuracy (MQ) & 1.00 & 1.00 & 1.00 & 1.00 & 1.00 & 1.00 & 1.00 \\
		Accuracy (TQ) & 1.00 & 1.00 & 1.00 & 1.00 & 1.00 & 1.00 & 1.00 \\
		Token $\Delta$ (\%) & 1.32 & 1.98 & 2.24 & 3.08 & 3.54 & 3.88 & 4.27 \\
		\hline
	\end{tabular}
\end{table}

\textbf{Analysis of Rendering Density Control.} We verify \textsc{Loom}'s ability to adhere to strict constraints by independently sweeping the total quota ($\Omega_{\text{total}}$) and per-verb caps ($\Omega_{\text{max}}$). First, regarding \textit{Total Quota Sensitivity}, we varied the scene-level budget $\Omega_{\text{total}} \in \{1, \dots, 7\}$ with fixed high per-channel caps. Table \ref{tab:quota_total} shows perfect adherence ($1.00$) for both Max-Quota (MQ) and Total-Quota (TQ) metrics. 

We also observe the generation overhead via Token $\Delta$, defined as the percentage length increase relative to the input. The token length increase scales linearly with the quota, confirming that \textsc{Loom} avoids uncontrolled verbosity.

\begin{table}[h]
	\centering
	\caption{Control accuracy under varying \textbf{Per-Verb Caps} ($\Omega_{\text{max}}$).}
	\label{tab:quota_channel}
	\renewcommand{\arraystretch}{1.15}
	\setlength{\tabcolsep}{3.5pt}
	\begin{tabular}{l|ccccccc}
		\hline
		\textbf{Cap Level} & \textbf{1} & \textbf{2} & \textbf{3} & \textbf{4} & \textbf{5} & \textbf{6} & \textbf{7} \\
		\hline
		Accuracy (MQ) & 1.00 & 1.00 & 1.00 & 1.00 & 1.00 & 1.00 & 1.00 \\
		Accuracy (TQ) & 1.00 & 1.00 & 1.00 & 1.00 & 1.00 & 1.00 & 1.00 \\
		Token $\Delta$ (\%) & 4.42 & 4.43 & 4.52 & 4.62 & 4.70 & 4.79 & 4.75 \\
		\hline
	\end{tabular}
\end{table}

Second, regarding \textit{Per-Verb Caps Sensitivity}, we varied the maximum visual quota $\Omega_{\text{max}} \in \{1, \dots, 7\}$ with a fixed total budget. Table \ref{tab:quota_channel} confirms perfect adherence ($1.00$) to local limits. The stable token expansion rate ($\approx 4.5\%$) indicates that the system successfully reallocates the fixed budget across unrestricted channels, proving it can throttle specific modalities without destabilizing overall generation.

\section{CONCLUSION}
This work introduces \textsc{Loom}, an agentic framework fundamentally focused on controllable narrative rendering. Moving past conventional remedial polishing, \textsc{Loom} operationalizes the distinction between story and discourse via a unique three-layer pipeline. This architectural approach ensures precise control over descriptive intensity while successfully preventing uncontrolled plot expansion. Our findings confirm the viability of integrating structural constraints for reliable, creative human-AI co-writing.

\section{Discussion and Limitations}
While validated on narrative texts, \textsc{Loom}'s reliance on linear story-discourse structures limits applicability to poetry and other specific literary genres. Furthermore, the dependency on three-layer prompt engineering hinders the handling of nonlinear plots. Future work will thus investigate agent-based automation and distillation to improve robustness and extend support to diverse genres.

\bibliographystyle{IEEEtran}   %
\bibliography{ref}            

\end{document}